\pdfoutput=1
\documentclass[11pt]{article}
\usepackage{authblk}
\usepackage{EMNLP2023}
\usepackage{times}
\usepackage{latexsym}
\usepackage[T1]{fontenc}
\usepackage[utf8]{inputenc}
\usepackage{microtype}
\usepackage{inconsolata}
\usepackage{booktabs}
\usepackage{graphicx}
\usepackage{graphics}
\usepackage[caption=false]{subfig}
\usepackage{hyperref}
\usepackage{tikz}
\usepackage{pgfplots}
\usepackage{todonotes}
\usepackage{amsmath}
\usepackage{cleveref}
\usepackage{array}
\usepackage{geometry}
\usepackage{pdflscape}
\DeclareMathOperator{\topk}{topk}
\usepackage{caption}
\newcounter{AlgoTableCount}
\newcounter{TempTableCount}
\newenvironment{algoTableCaption}{%
    
    \setcounter{TempTableCount}{\thetable}
    \setcounter{table}{\theAlgoTableCount}
    }{%
    \setcounter{table}{\theTempTableCount}
    \stepcounter{AlgoTableCount}
    }

\graphicspath{{./figures/}}

\title{Arctic-Embed: Scalable, Efficient, and Accurate Text Embedding Models}
\author[1,*]{Luke Merrick}
\author[1]{Danmei Xu}
\author[1]{Gaurav Nuti}
\author[1]{Daniel Campos}
\affil[1]{Snowflake Inc. }
\affil[*]{\small Corresponding author, \texttt{luke.merrick@snowflake.com}}

\begin{document}
\maketitle

\begin{abstract}
This report describes the training dataset creation and recipe behind the family of \texttt{arctic-embed} text embedding models (a set of five models ranging from 22 to 334 million parameters with weights open-sourced under an Apache-2 license). At the time of their release, each model achieved state-of-the-art retrieval accuracy for models of their size on the MTEB Retrieval leaderboard,\footnote{\url{https://huggingface.co/spaces/mteb/leaderboard}} with the largest model, arctic-embed-l outperforming closed source embedding models such as Cohere's embed-v3 and Open AI's text-embed-3-large. In addition to the details of our training recipe, we have provided several informative ablation studies, which we believe are the cause of our model performance.
\end{abstract}

\section{Introduction}
Embedding models' ability to provide accurate retrieval performance without additional tuning \cite{lewis2020retrieval} has made them a popular choice in search and Retrieval-augmented-generation (RAG) \cite{Ram2023InContextRL} workloads. \\
Unlike traditional keyword search, embedding models encode information beyond token overlap. This allows embedding systems to represent queries like \textit{How tall is Tom Cruise?} and \textit{Height of the actor who plays Maverick in Top Gun} closely despite having no common words.

\begin{figure}
    \centering
    \includegraphics[width=\linewidth]{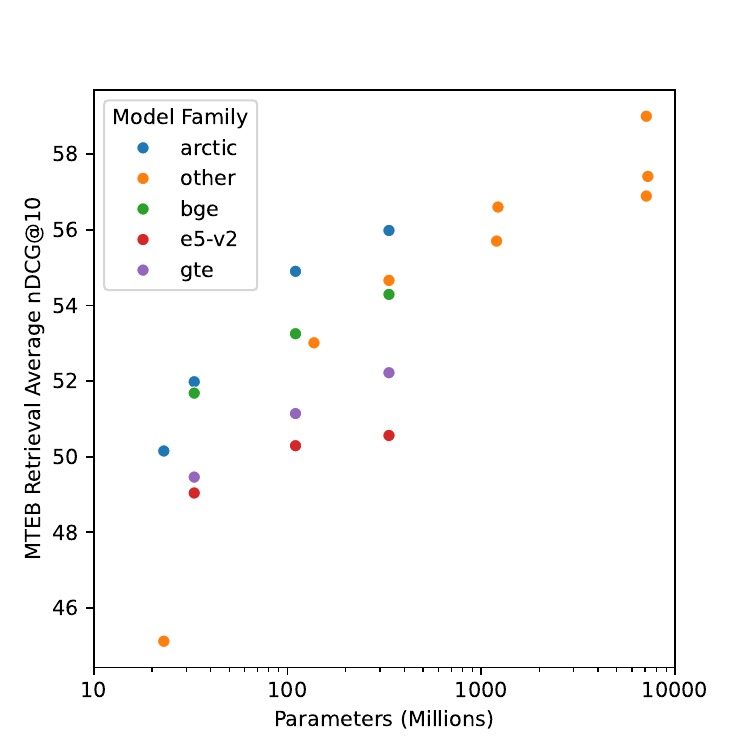}
    \caption{Snowflake's Arctic-embed models are a suite of 5 embedding models, each of which pushes the Pareto frontier in the trade-off between model size and retrieval performance on the MTEB Retrieval Leaderboard.}
    \label{fig:tradeoff}
\end{figure}

Driven by the utility and the widespread adoption of these models, the broader open-source and research community has put forth a constant stream of ever-stronger text embedding models such as E5 \cite{Wang2022TextEB}, GTE \cite{Li2023TowardsGT}, Jina \cite{günther2024jina}. The quick experimentation and improvement underpinning these works is, in turn, thanks in part to large-scale open evaluation benchmarks such as MSMARCO \cite{Campos2016MSMA}, BEIR \cite{thakur2021beir}, and MTEB \cite{muennighoff2023mteb}. These leaderboards combine an easy and efficient evaluation with a broad array of tasks, which allows for effective experimentation.

This paper's work was motivated in early 2024 by the lack of efficient and effective open-text embedding models competing with the performance of closed-source models such as Cohere's \texttt{embed-v3} or OpenAI's \texttt{text-embed-3-large}. While models such as 
SFR-Embedding-Mistral \cite{SFRAIResearch2024} and GritLM \cite{muennighoff2024generative} outscore proprietary offerings; their size (each over 7 billion parameters) and their dimensionality (each 4096) make them impractical to use in many production workloads. Seeking to provide a high-quality retrieval model with fewer than a billion parameters, we set out to train a suite of high-quality embedding models.

Through fruitful data-centric experiments, we developed the recently released Arctic family of text embedding models. Based on five encoder-only pretrained language models of various sizes (see \Cref{tab:model-sizes}) and leveraging the same training data and methodology, we trained each model to optimize retrieval performance as measured by nDCG@10 on the MTEB Retrieval leaderboard. As shown in \Cref{fig:tradeoff}, each variant achieved a new state-of-the-art performance for its size.\footnote{As of April 16th, 2024}. We present these models and this technical report as a journal of our experiments that led to our improvements in performance.

\subsection{Summary of Contributions}

\noindent\textbf{Open model release.} We release a suite of embedding models, Arctic-embed, under a permissive Apache-2 license, which delivers state-of-the-art retrieval performance for their size/context window class on the Retrieval portion of the MTEB leaderboard.\\

\noindent\textbf{Demonstrated importance of data organization.} We present a set of ablations that suggest improvements in retrieval quality are more strongly tied to data sampling during training and the method of negative mining than scaling up data scale and batch size, where previous work has focused.\\

\noindent\textbf{Improved methods for synthetic data.} We present a novel technique for query generation grounded by mined hard negatives, which we found more effective than straightforward generation approaches that generate both queries and negatives and which served as a key ingredient in our models' success. 
\begin{table*}
    \centering
    \small
    \begin{tabular}{llll}
        \toprule
         Size&  Base Model (Huggingface ID) &  Parameters (M) & Embedding Dimension\\
         \midrule
         \texttt{xs} & \texttt{nreimers/MiniLM-L6-H384-uncased} & 23 & 384\\
         \texttt{s} & \texttt{intfloat/e5-unsupervised-small} & 33 & 384\\
         \texttt{m} & \texttt{intfloat/e5-unsupervised-base} & 110 & 764\\
         \texttt{m-long} & \texttt{nomic-ai/nomic-embed-text-v1-unsupervised}& 137 & 768\\
         \texttt{l} & \texttt{intfloat/e5-unsupervised-large} & 334 & 1024\\
    \end{tabular}
    \caption{Breakdown of model architectures.}
    \label{tab:model-sizes}
\end{table*}
\section{Background}
\subsection{Task Description}
An embedding model maps a variable input into a fixed-dimensional vector. This one-way transform can be applied to various modalities, scales, and scopes and directly used for downstream tasks such as classification, clustering, or retrieval. In the scope of our work, we focus on text embeddings for retrieval. This task aims to train a model that maximizes the similarity between relevant documents, given a query and a document collection, while minimizing the similarity with irrelevant documents.\\

The representation-based retrieval method has emerged as a standard paradigm as it minimizes the frequency with which inputs are transformed into vectors. Offline, the document corpus is processed, resulting in a set of vectors stored in an Approximate Nearest Neighbor Index such as FAISS \cite{douze2024faiss}. The input is transformed online into a vector at query time, and the documents with the closest embeddings are retrieved. In other words, the cosine distance between queries and documents signals relevance. 

\subsection{Related Work}
\textbf{Training Approaches}: Building on prior success in NLP and IR research knowledge, embedding model training uses supervised learning with examples of positive and negative document query pairs. It is common to use labeled positive and negative query-document retrieval examples (commonly extracted from weak signals or labeled data \cite{Lin2020PretrainedTF}) to fine-tune general-purpose pre-trained language models into specialized text embedding models. In this paradigm, \citet{qu2021rocketqa} demonstrated the importance of scaling the batch size and training on hard negatives. In contrast, \citet{Xiong2020ApproximateNN} demonstrated the importance of adapting the negatives' difficulty to the retriever's competence.

While earlier work focused on leveraging supervised datasets such as HotpotQA \cite{Yang2018HotpotQAAD} or NQ \cite{Kwiatkowski2019NaturalQA}, \citet{Wang2022TextEB} demonstrated the effectiveness of constructing large datasets from web-crawled title-document examples through the groundbreaking performance of their resulting E5 model. \citet{bge_embedding} and \citet{nussbaum2024nomic} combine generated datasets with supervised labeled datasets to improve retrieval performance further. 

\textbf{Model Architecture}: Building on the success and utility of the transformer \cite{Vaswani2017AttentionIA} prior work has primarily focused on training models using BERT \cite{devlin2018bert}, or variants thereof. While some work has studied the usage of sequence to sequence \cite{zhuang2022rankt5} or large decoder-only models \cite{SFRAIResearch2024} \cite{muennighoff2024generative}, these models' increased model size and associated worse inference efficiency have kept the majority of focus on encoder-only variants.

\textbf{Training Objective}: Many works initially trained retrievers and rankers leveraging traditional loss forms such as Mean Squared Error \cite{Lin2020PretrainedTF}. Still, recently, the application of a contrastive loss \cite{Hadsell2006DimensionalityRB, Mueller2016SiameseRA}, which leverages not only positive pairs but the relationship between positive and negative pairs, has risen to prominence. InfoNCE (Noise Contrastive Estimation) \cite{Oord2018RepresentationLW} improved on the constrastive triplet loss and has quickly become one of the most popular and common losses used to train embedding models.

\section{Arctic Embed}

\begin{table*}
    \centering
    \small
    % NOTE: The ">{\raggedright\arraybackslash}" thing disables word hyphenation for this column, see https://tex.stackexchange.com/a/2892. 
    \begin{tabular}{p{0.23\linewidth}p{0.5\linewidth}>{\raggedright\arraybackslash}p{0.15\linewidth}}
        \toprule
         Difference &  Description &  Ablation Study \\
         \midrule
         % Difference
         Better data &
         % Description
         We leverage web search data and common web data filtering methods.&
         % Ablation
         Shows Improvement \\

         % Difference
         Source stratification &
         % Description
         Discussed further in Section~\ref{sec:stratify}, we fill each minibatch of training data with examples from a single data source. Nomic also uses the technique. &
         % Ablation
         Shows Improvement \\

         % Difference
         Longer pretraining sequence length &
         % Description
         We used a maximum sequence length of 256, which is twice as long as in GTE and BGE \cite{li2023general, xiao2023cpack}, despite using approximately the same batch size as these works.&
         % Ablation
         Shows Improvement \\

         % Difference
         Base models trained for retrieval &
         % Description
         Several sizes of Arctic-embed use pre-trained text embedding models as starting points. For example, we use \texttt{e5-unsupervised-base} instead of its parent model \texttt{bert-base-uncased} as the backbone of \texttt{arctic-embed-m}.&
         % Ablation
         Shows Inconsistent Improvement \\

         % Difference
         \texttt{[CLS]} embeddings &
         % Description
         Unlike most prior art, Arctic embed uses \texttt{[CLS]} token embeddings instead of mean pooling. BGE also uses the technique. &
         % Ablation
         Not Studied \\

         % Difference
         Implementation and tuning &
         % Description
         We iterated relentlessly on the data mix, negative mining strategy, batch size, and other training parameters to double down on our data advantages in both training rounds. Additionally, our training implementation includes several attention-to-detail tricks (e.g., in-batch document deduplication), which may improve performance compared to more naive implementations.&
         % Ablation
         Not Studied
    \end{tabular}
    \caption{Differences from prior works hypothesized to help Arctic embed score higher on MTEB Retrieval.}
    \label{tab:comparison}
\end{table*}

With Arctic-embed, we aimed to start from the current consensus of best practices from the literature and train an embedding model from the ground up. 

Consistent with prior works, like E5, BGE, GTE, Jina, and Nomic \cite{wang2024e5, xiao2023cpack, li2023general, günther2024jina, nussbaum2024nomic}, we conduct two training rounds using two different kinds of datasets. The initial training round is large-scale pretraining using only in-batch negative examples. This round of training leverages a dataset of pairs of queries and relevant documents. The second round of training (often referred to as the fine-tuning step) calls for similar pairs of queries and documents augmented with an additional set of ``hard'' negative documents (where ``hard'' refers to the fact that it is not trivial to determine their lower relevance relative to the labeled-as-relevant document). We used a tunable negative mining strategy (see \Cref{sec:training-data}) to construct a focused dataset of about a million samples for this round of training.

Although our work closely replicates many of the steps prior works took, our resulting models score higher on the MTEB Retrieval benchmark, sometimes by a substantial margin. In \Cref{tab:comparison} we present several hypotheses about what led to this improved performance, and in \Cref{sec:ablation} we test several of these hypotheses through ablation studies.

\subsection{Model Architecture}
We trained models of varying sizes from BERT-like backbones as shown in \Cref{tab:model-sizes}. Our \texttt{m} and \texttt{l} are standard BERT architecture \cite{devlin2019bert} (BERT \texttt{base} and \texttt{large}, respectively). We looked to variants of the MiniLMv2 architecture \cite{wang2021minilmv2} for our smaller sizes (\texttt{xs} and \texttt{s}), and we opted for the Nomic BERT architecture \cite{nussbaum2024nomic} for our long-context variant (\texttt{m-long}).

\subsection{Pooling}
Architecturally, we do not modify any base model, even just the common practice of adding a pooling layer to the base model.\footnote{e.g., we use \texttt{AutoModel.from\_pretrained(\ldots, add\_pooling\_layer=False)} in the \texttt{transformers} Python package)} Additionally, instead of pooling output vectors, we utilize the final hidden state of the \texttt{[CLS]} token as the embedding vector, in contrast to the mean pooling strategy used in E5, GTE, and Nomic \cite{wang2024e5, li2023general, nussbaum2024nomic}. This choice matches the BGE architecture \cite{xiao2023cpack} and is inspired by the ablation study in \cite{li2023angleoptimized}, showing this led to a 2.5\% higher score on the Semantic Text Similarity (STS) evaluation studied. 

\subsection{Training Data}\label{sec:training-data}
In creating our training datasets, we took inspiration from the world of Large Language Models (LLMs) and leveraged filtering methods inspired by RefinedWeb \cite{penedo2023refinedweb}, C4 \cite{rae2022scaling}, Gopher \cite{raffel2023exploring}, and TogetherAI \cite{together2023redpajama}. 

First, for noisy raw data sources such as web search, we parse structured web documents using trafilatura\footnote{\url{https://trafilatura.readthedocs.io/en/latest/}}. While parsing, we compute custom signals for quality filtering. Specifically for positive data pair cleaning, we need to ensure: a) each text in the pair is of good quality (language filter, text quality filter) and b) text pairs (query, document) are similar in meaning (consistency filter). For quality filtering, we leverage a series of filters similar to ones detailed in Snowflake's Arctic model training cookbook\footnote{\href{https://medium.com/snowflake/snowflake-arctic-cookbook-series-arctics-approach-to-data-b81a8a0958bd}{https://medium.com/snowflake/snowflake-arctic-cookbook-series-arctics-approach-to-data-b81a8a0958bd}}. A complete list of effective filtering methods can be found in \Cref{sec:appendix-filter-list}.
We combine these filters to create a more curated dataset by removing low-quality, irrelevant, or potentially spam documents based on various characteristics related to content quality, language structure, and duplication.

For consistency filtering, we apply a low-fidelity, high-throughput pair-similarity consistency filter — sentence similarity using a \texttt{fastText}\footnote{\url{https://fasttext.cc/docs/en/english-vectors.html}} word2vec model (which can be run cheaply on CPU). Rather than treating these embeddings' signal as a clear quality label, we instead adopt a conservative threshold (a low minimum allowed similarity of 0.3) and use them to filter out unrelated examples (e.g., ``CGplayer doesn't work properly without JavaScript-enabled'' documents from web crawl failures). Additionally, we truncate long sequences to 512 words during this step. As we observed, queries in the web-based corpus were usually answered at the beginning of the document. Not only was it computationally wasteful, but even the meaning captured in word2vec embeddings would get diluted by averaging vectors from irrelevant words present later.

\subsection{Dataset Mix And Sampling}
Due to the different datasets' sizes, consistency, hardness, and learning dynamics, simply concatenating all available datasets together proved a suboptimal strategy, especially in the fine-tuning stage. Instead, we ran isolated experiments to understand the effects of each dataset on fine-tuned performance. Then, we selected and combined datasets based on their relative performance in these experiments. Each data source we used is described in more depth below.

\begin{figure}
    \centering
    \includegraphics[width=1.0\linewidth]{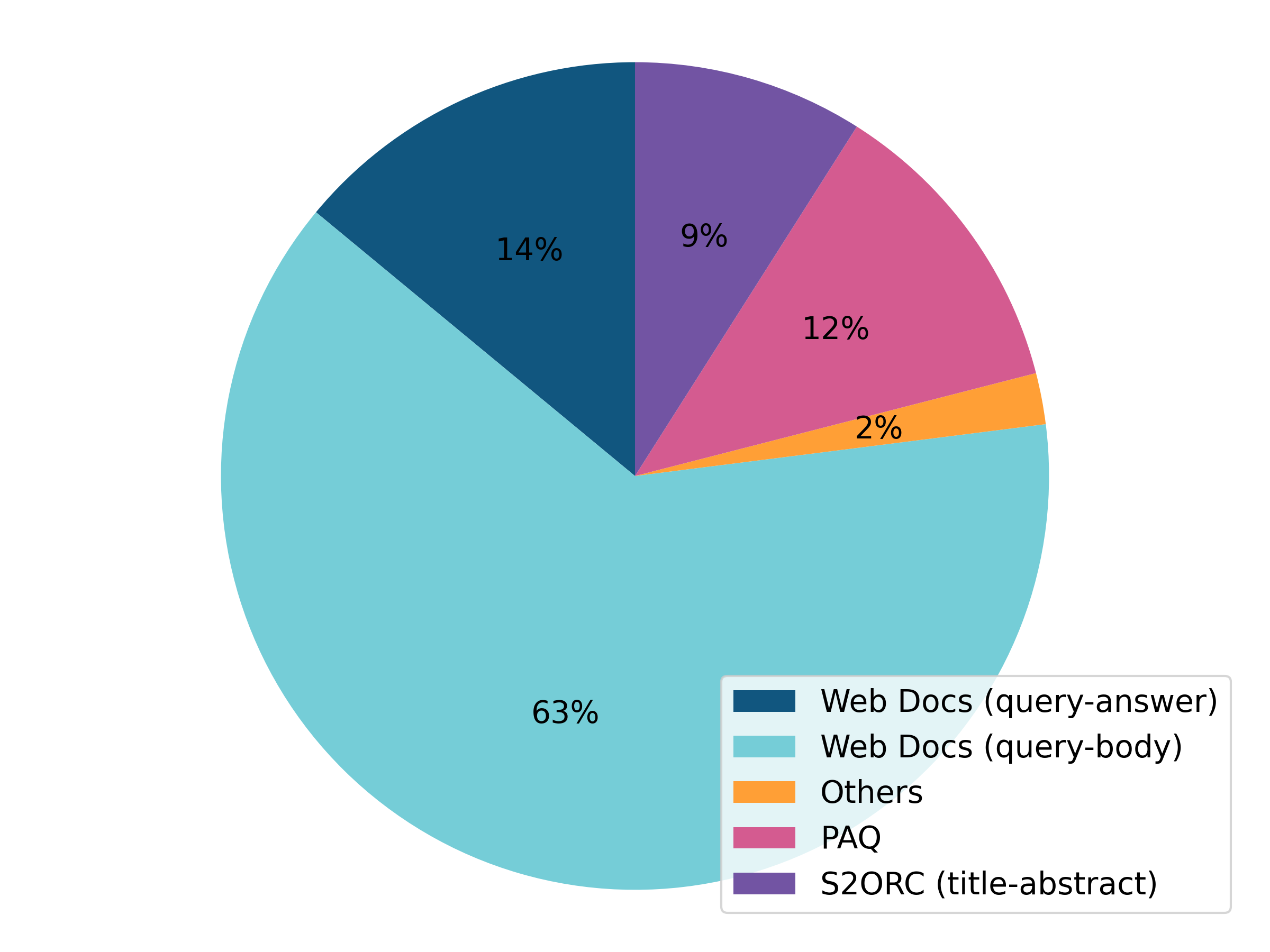}
    \caption{Composition of our pretraining dataset containing about 300 million query document pairs.}
    \label{fig:pt-mix}
\end{figure}

\begin{figure}
    \centering
    \includegraphics[width=1.0\linewidth]{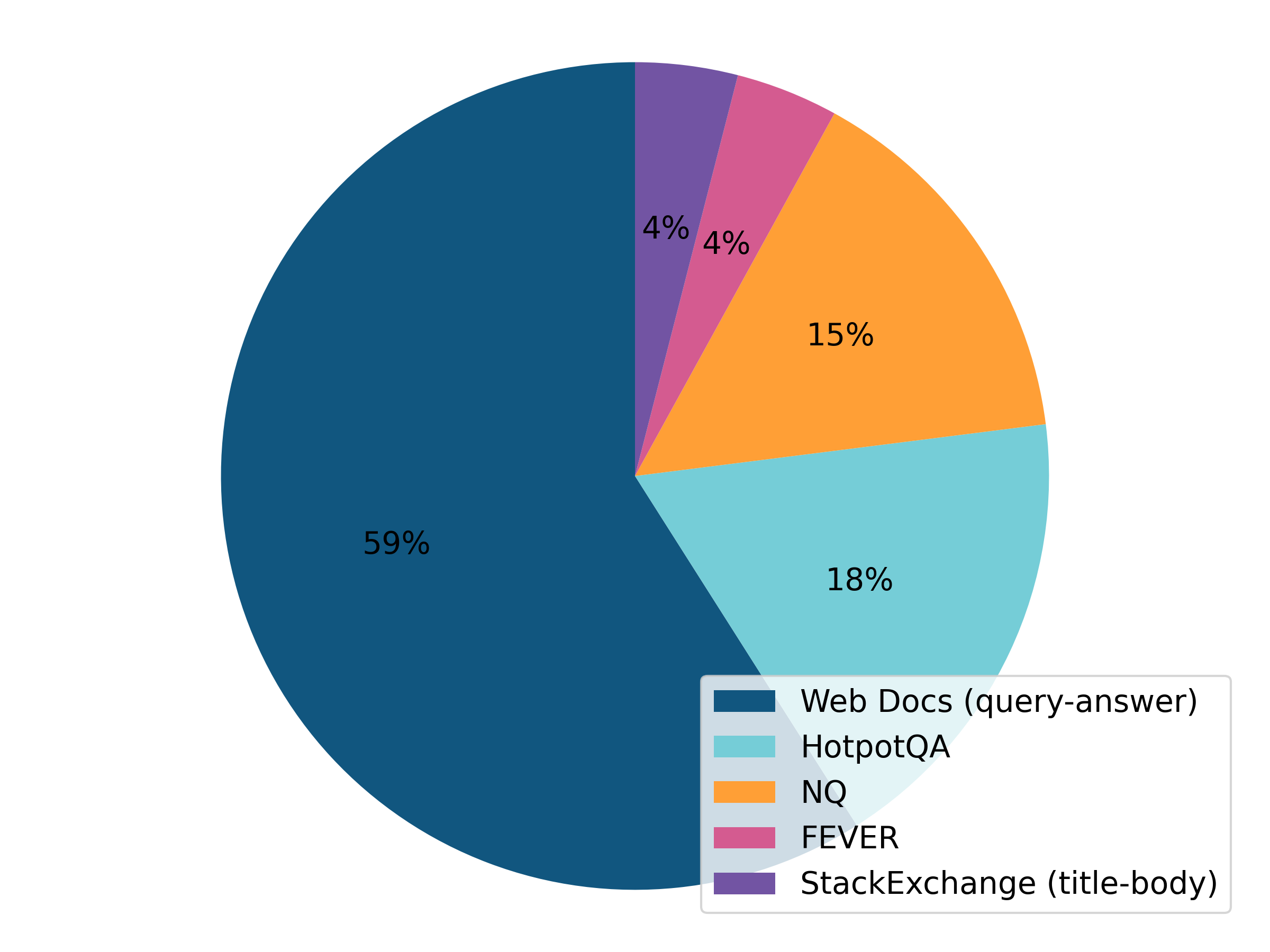}
    \caption{Composition of our finetuning dataset, which contains about 1 million queries, each paired with a positive document and a set of hard negative documents.}
    \label{fig:ft-mix}
\end{figure}

Our large pretraining dataset, described in figure \ref{fig:pt-mix}, amounts to 308 million query-document pairs (filtered from around 2 billion documents), of which 71\%  are web search documents paired with either a query or title. Aside from web search data, text pairs set include PAQ\footnote{\url{https://github.com/facebookresearch/PAQ}}, StackExchange title-body and title-body web documents pairs from common crawl based sources, and S2ORC title-abstract pairs\footnote{\url{https://github.com/allenai/s2orc}}. We have found the previous steps on quality annotation and filter transformative in improving quality and pruning noise in web search data and beyond for pairwise positive datasets.

Our fine-tuning dataset, described in \Cref{fig:ft-mix}, consists of around 1 million pairs built by combining our web search data with several public datasets (HotpotQA\footnote{\url{https://github.com/hotpotqa/hotpot}}, NQ\footnote{\url{https://github.com/google-research-datasets/natural-questions}}, Fever\footnote{\url{https://fever.ai/dataset/fever.html}}, and StackExchange title-body\footnote{\url{https://huggingface.co/datasets/sentence-transformers/embedding-training-data}}), and then performing further expansion via synthetic mining strategy detailed in section below. This mix notably omits several popular public datasets used by other embedding models due to our observation of positive pair consistency and negative pair level of hardness. These found-to-be-less-useful datasets include NLI, MEDI, WikiAnswers, and SQuAD. Empirically, we have observed that quantity is less important than quality in the finetuning phase, and an overpowering amount of low-quality data can lead to lower-quality models.

\subsection{Synthetic Data For Semantic Dense Mining}

\begin{figure}
    \centering
    \includegraphics[width=1.0\linewidth]{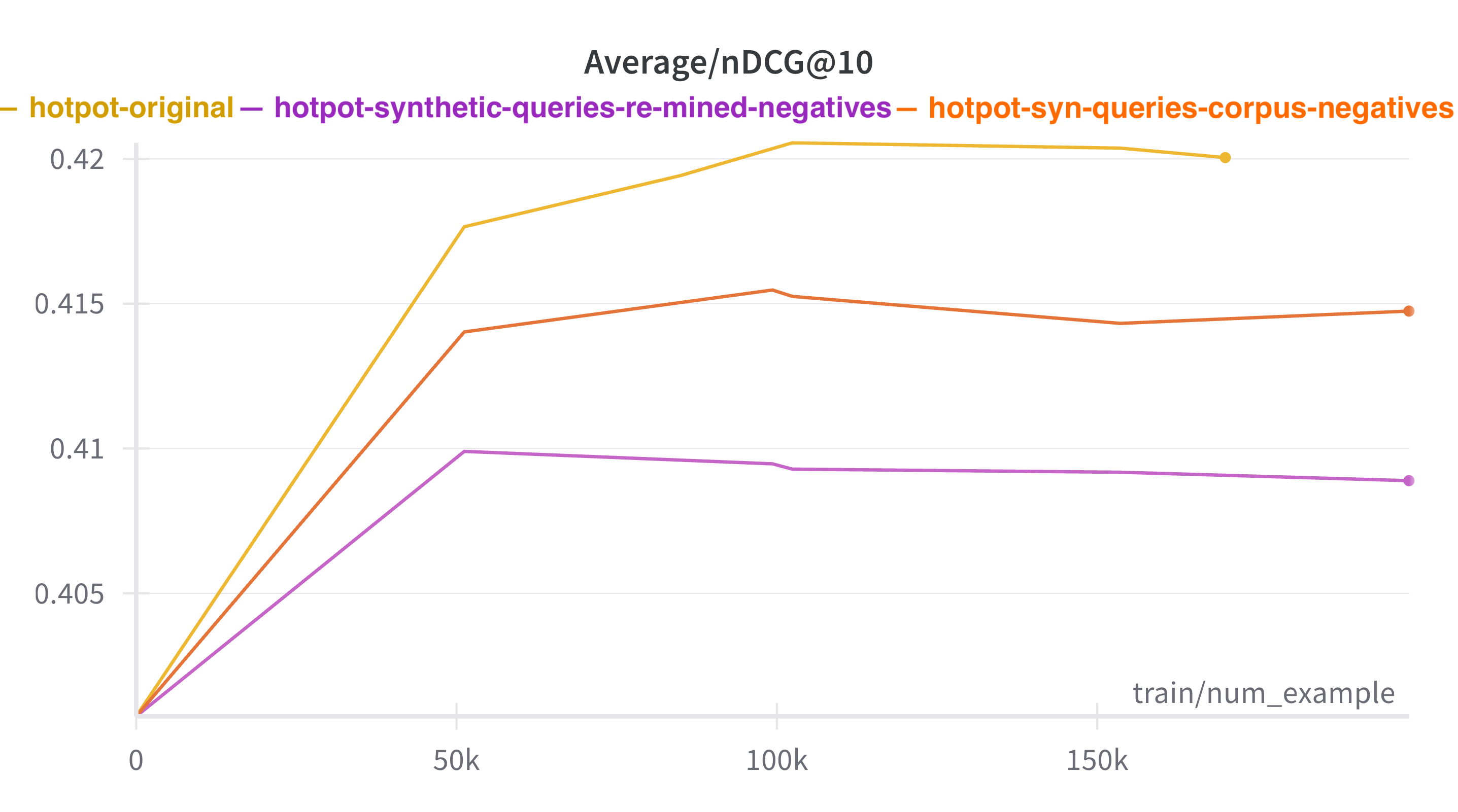}
    \caption{Comparison of model performance when training using synthetic datasets generated from the HotpotQA document corpus versus the original HotpotQA queries (using Algorithm~\ref{algo:negative-mine} for negative mining in both cases).}
    \label{fig:hotpotqa-synthetic}
\end{figure}

Compared to the abundance of web-scale data used in pretraining, high-quality examples suitable for finetuning are more scarce. To address this data scarcity, we used synthetic data creation to construct additional datasets that benefited downstream performance just as much as those listed above. Similar to the prior work of \citet{dai2022promptagator,lee2024gecko}, we leverage Large Language Models to generate novel queries. Breaking from these previous approaches, however, we found it critical to add negative documents to our LLM inputs to ground the query generation (see Algorithm~\ref{algo:synthetic} in the Appendix for details). Additionally, we chose to generate only synthetic queries rather than synthetic negatives because we found that LLMs do not easily generate relevant negatives of as high quality as those mined from a preexisting corpus of documents. \Cref{fig:hotpotqa-synthetic} shows this approach in action -- two datasets generated by variants of Algorithm~\ref{algo:synthetic} led to score increases approaching that afforded by the original HotpotQA.

\subsection{Tunable Hard Negative Mining}
\label{sec:optimize}
Fine-tuning datasets typically include carefully chosen ``hard'' negative examples and a positively-relevant query-document pair. How hard should these negatives be for maximally effective learning in the fine-tuning phase? Our answer to this question was ultimately a \textit{tunable hard negative mining} strategy in which we leveraged a preexisting text embedding model to identify and score the hardest negatives for each training example. Then, we applied a score threshold to discard the hard negatives from the above set. We found that using an upper threshold rather than a specific rank helped account for the fact that some queries admit much harder top-k negatives than others, and in \Cref{sec:fine-tune-ablation}, we perform a parameter sweep of the negative hardness threshold to demonstrate the value of a tunable approach (the optimal threshold value scores significantly better than other choices). We additionally note that although Algorithm~\ref{algo:negative-mine} indicates both an upper and lower relevance threshold for negative mining, in practice, we retrieved the top 100 hardest negatives and applied only an upper threshold as a performance optimization.

\begin{figure}
    \centering
    \includegraphics[width=1.0\linewidth]{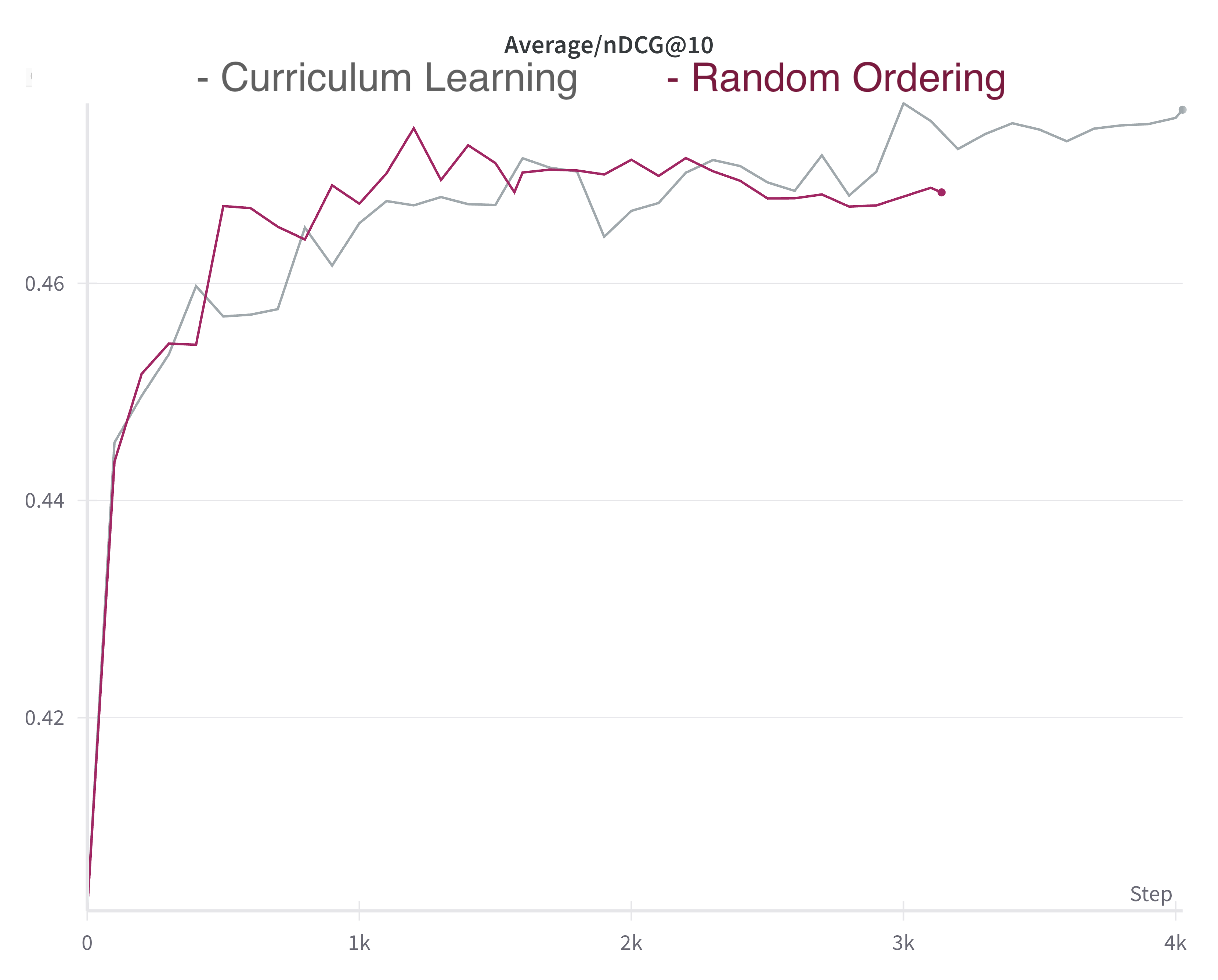}
    \caption{Impact of curriculum learning on nDCG@10 retrieval score during training. Scores averaged over a handful of small datasets used for fast in-training evaluation.}
    \label{fig:ft-threshold-currciulum}
\end{figure}

Beyond tuning to a single hardness threshold level, we hypothesized that ordering the data by the difficulty of the negatives (i.e., curriculum learning) could lead to even better results. In this vein, we offer the experiment shown in \Cref{fig:ft-threshold-currciulum}, which compares the Impact of training with negatives of progressively increasing difficulty. While this initial experiment suggests some improvement in curating the curriculum of hard negatives, we note that this experiment was run after the release of the Arctic embed, and we did not use this curriculum approach when training our published models.

\section{Training Recipe}
\begin{table}
    \centering
    \tiny
    \begin{tabular}{lllll}
        \toprule
         Variant & Pre Batch & Pre LR & Finetune Batch & Fine LR \\
         \midrule
         \texttt{xs} & 24,576 & 6e-4& 768 & 4e-5 \\
         \texttt{s} & 32,768 & 5e-4 & 1,024 & 4e-5 \\
         \texttt{m} & 16,384 & 2e-4 & 512 & 1e-5 \\
         \texttt{m-long} & 12,288 & 1e-4 & 512 & 1e-5 \\
         \texttt{l} & 12,480  & 1e-4 & 512 & 9e-6 \\
    \end{tabular}
    \caption{Learning rate and batch size for both rounds of training.}
    \label{tab:hyperparams}
\end{table}

\subsection{Model Initialization}
\begin{figure}
    \centering
    \includegraphics[width=1.0\linewidth]{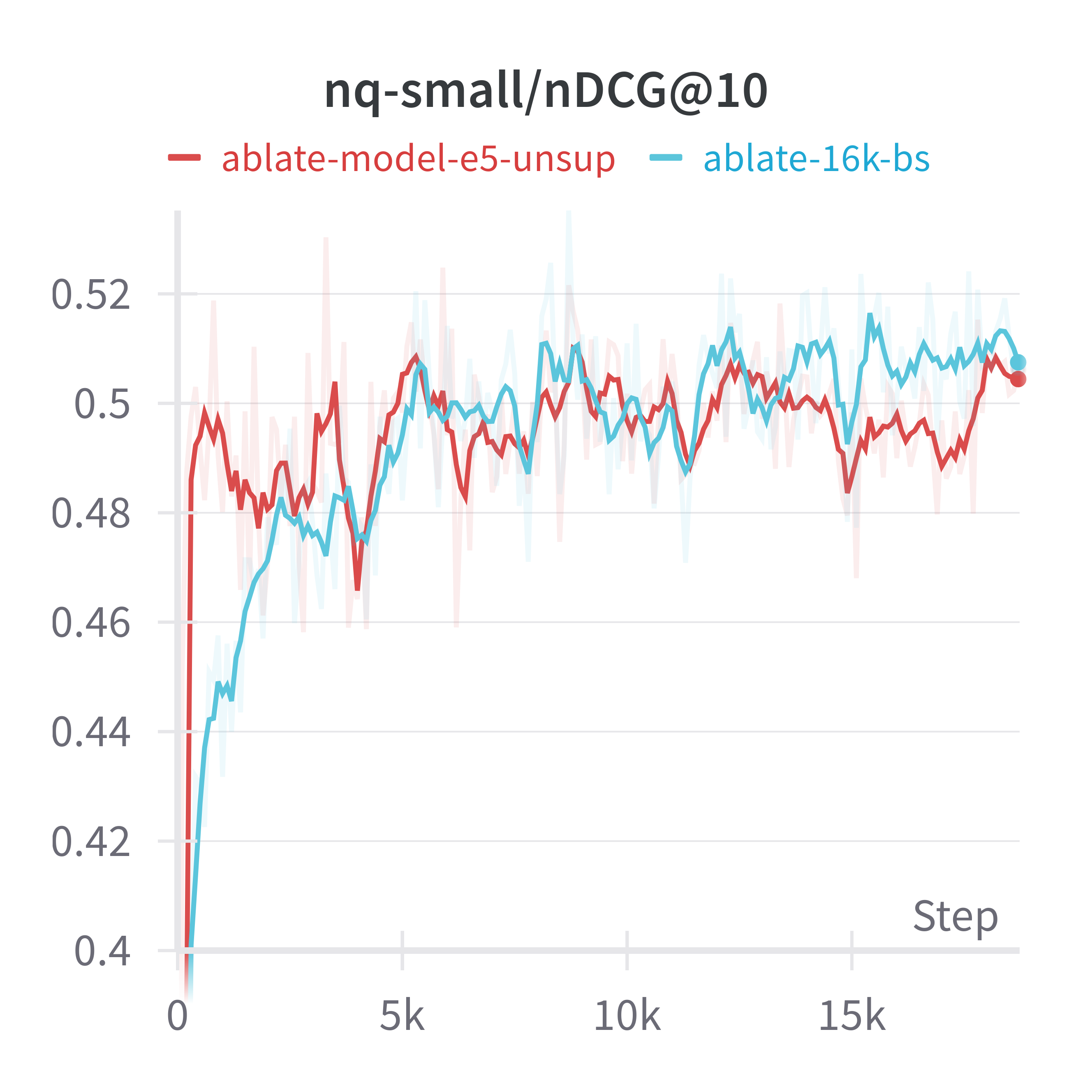}
    \caption{A granular look at our starting weight ablation study from Section~\ref{sec:ablation-pretrain}. We plotted the rolling average nDCG@10 score throughout training, observing that the run using pre-trained E5 model weights (red) converged much more quickly than the run using general-purpose BERT weights (blue). The evaluation dataset is a ``lite BEIR'' dataset based on NQ -- details in \Cref{sec:lite}.}
    \label{fig:base-training-curve}
\end{figure}

We begin with a pretrained language model. Where permissively licensed base models pre-trained for information retrieval are available for a given model size, we prefer these weights over general-purpose pretrained ones \footnote{Such as https://huggingface.co/intfloat/e5-large-unsupervised}. Our ablation studies in \Cref{sec:ablation-pretrain,sec:ablation-e2e} showed mixed results and suggested the effect of this design choice on performance may have been weak relative to other effects studied, but as \Cref{fig:base-training-curve} shows, starting from a more thoroughly trained base model, such as \texttt{e5-base-unsupervised}, had an apparent effect on sample-efficiency and convergence speed, and this speedup was notably helpful for faster experimentation during model development.

\subsection{Large Scale Contrastive Pretraining With In-Batch Negatives}
In the first round of contrastive training, we aim for large scale, both in batch and total dataset sizes. We use our pretraining dataset with the infoNCE contrastive loss using in-batch negatives (for each query, all documents associated with different queries in the minibatch are treated as negative examples). GPU parallelism, activation checkpointing, and truncated sequence length were instrumental in achieving large batch sizes.

We train for one epoch\footnote{Some sizes of Arctic embed utilized early checkpoints from before one epoch of pretraining, though this was done for expediency, and we did not find evidence of this improved performance.} using the AdamW optimizer, adjusting only the learning rate while leaving all other parameters at PyTorch default values. We perform a linear learning rate warmup for several hundred steps, then a linear decay to 10\% of the original learning rate over the remainder of the training. As evidenced by the example shown in \Cref{fig:ablate-schedule}, we observed performance could be sensitive to the learning rate and the learning rate schedule. Batch sizes and learning rates for each model size are given in \Cref{tab:hyperparams}.

\subsection{Longer Truncation Length}
Much of our pretraining data included documents significantly longer than 128 tokens. We used a document sequence length of 256 in large-scale contrastive training, in contrast to the 128 truncation length used in GTE and BGE. We truncated query sequence length to 32, consistent with BGE's source code\footnote{\url{https://github.com/FlagOpen/FlagEmbedding/blob/53cfac4a50ac0e023b2f8d19b10667c9c210fa41/FlagEmbedding/baai_general_embedding/finetune/arguments.py}}. Our ablation study in \Cref{sec:ablation} suggests this longer truncation length led to a substantial improvement in retrieval performance.

\subsection{Source Stratification}\label{sec:stratify}
We fill each batch with data from a single source during pretraining, a source of accuracy gains in prior work \cite{nussbaum2024nomic}. Our ablation study in \Cref{sec:ablation-pretrain} indicates this led to a dramatic improvement in model quality (see \Cref{tab:pretrain-ablate}).

\subsection{Quality-Focused Contrastive Training With Curated Negatives}
After large-scale training, we perform a second round of training leveraging our fine-tuning dataset, which contains explicitly labeled negative examples. We use no learning rate warmup but apply the same linear learning rate decay schedule as in the pretraining stage. We truncate sequence lengths to 512 for queries and documents for all models, including the long-context variant \texttt{m-long}. For each query in a batch, we include one positive document and ten hard negative documents. Batch sizes (number of queries) and learning rates for each model size are given in \Cref{tab:hyperparams}.

\subsection{Disabling In-Batch Negative Loss}
Based on some early fine-tuning runs, we found that disabling in-batch negative loss did not measurably degrade performance. We stopped using in-batch negatives during fine-tuning (this made tuning easier, especially since the interaction between batch size and in-batch loss is not straightforward). 

\section{Efficiency}\label{sec:efficiency}
To maximize experimental throughput, iteration speed, and the maximum feasible batch size, we took pains to ensure our training setup was as effective as possible for our given computational budget. We carefully optimized the net efficiency of training and iteration, assuming a single training node with 8 NVIDIA H100 GPUs. We achieved high efficiency by carefully implementing a custom data loader and writing our training loop in plain PyTorch to leverage several ``tricks'' we detail in \Cref{sec:appendix-efficiency-tricks}. Additionally, we identified and eliminated careless performance bottlenecks through performance benchmarking, keeping a watchful eye on both throughput and GPU utilization. 

Further discussions about methods we found helpful for efficient experimentation and training can be found in the Appendix, including a discussion of granular evaluation during training in \Cref{sec:lite}.

\section{Experimental Results}
To qualify our retrieval quality, we evaluate model performance on the Retrieval portion of the MTEB dataset \cite{muennighoff2023mteb}. Summary results of MTEB experiments are shown in \Cref{fig:tradeoff}, and a complete tabulation by dataset is given in \Cref{sec:appendix-full-table}. To quality the performance of our long context model, we leverage the LoCo \cite{saadfalcon2024benchmarking}, the results of which are given immediately below.

\subsection{Long Context Performance}

\begin{table*}
    \centering
    \small
    \scalebox{0.9}{
    \begin{tabular}{llllllll}
        \toprule
         Model & Seq Len & Summ. Scr. FD & Gov. Report & QMSUM & QASPER Title & QASPER Abs. & Avg \\
         \midrule
         \texttt{arctic-embed-m-long} & 2048 & 63.2 & 93.8 & 45.8 & 77.3 & 95.8 & 75.2 \\
         \texttt{arctic-embed-m-long} & 4096 & 81.8 & 96.0 & 39.7 & 85.9 & 99.0 & 80.5 \\
         \texttt{arctic-embed-m-long} & 8192 & 86.6 & 96.5 & 31.7  & 81.6 & 99.7 & 79.2 \\
        \midrule

        \texttt{jina-base-v2} & 2048 & 87.2 & 97.7 & 35.1 & 95.3 & 99.7 & 83.0 \\
        \texttt{jina-base-v2} & 8192 & 93.3 & 98.6 & 40.8 & 95.1 & 99.3 & 85.5 \\
        \texttt{nomic-embed-text-v1} & 2048 & 86.1 & 96.9 & 47.8 & 96.1 & 99.7 & 85.3 \\
        \texttt{nomic-embed-text-v1} & 4096 & 89.0 & 97.4 & 45.7 & 95.8 & 99.9 & 85.6 \\
        \texttt{nomic-embed-text-v1} & 8192 & 90.9 & 97.8 & 44.2 & 94.9 & 99.9 & 85.5 \\
        \texttt{e5-mistral} & 4096 & 95.9 & 98.3 & 46.8 & 98.4 & 99.8 & 87.8 \\
    \end{tabular}}
    \caption{Snowflake-arctic-embed-m-long nDCG@10 scores on the LoCo benchmark. Non-Arctic scores taken from \cite{nussbaum2024nomic} without an attempt to reproduce.}
    \label{tab:long-context}
\end{table*}

For our initial Arctic embed release, we did not put any special efforts into adjusting our training recipe for long-context support. Instead, our \texttt{m-long} variant was only trained on short sequence data (it was pretrained with sequences truncated to 256 tokens and finetuned with sequences truncated to 512 tokens). Nonetheless, even on the specialized LoCo long context benchmark datasets (\Cref{tab:long-context}), performance only tends to lag slightly compared to models trained end-to-end specifically with long-context in mind, e.g. \texttt{nomic-embed-text-v1}. While these LoCo results suggest \texttt{m-long} may not be the model of choice for long sequences, its strong MTEB Retrieval scores suggest it may be a good pick for datasets containing a mix of long and short sequences.

This surprisingly not-so-bad performance may be largely thanks to the base model of \texttt{m-long}, \texttt{nomic-embed-unsupervised}, being trained on long sequence retrieval, but unfortunately we did not have time to run an ablation study to quantify the Impact of this base model.\\

\section{Ablation Studies}\label{sec:ablation}
We conducted several ablation studies to test some of the hypotheses stated in \Cref{tab:comparison} regarding the causes of Arctic embed's higher MTEB Retrieval scores relative to other similar models. Average scores are given in the following subsections, with full scores in \Cref{sec:appendix-full-table}.

\subsection{Pre-training Ablations}\label{sec:ablation-pretrain}

\begin{figure}
    \centering
    \includegraphics[width=1.0\linewidth]{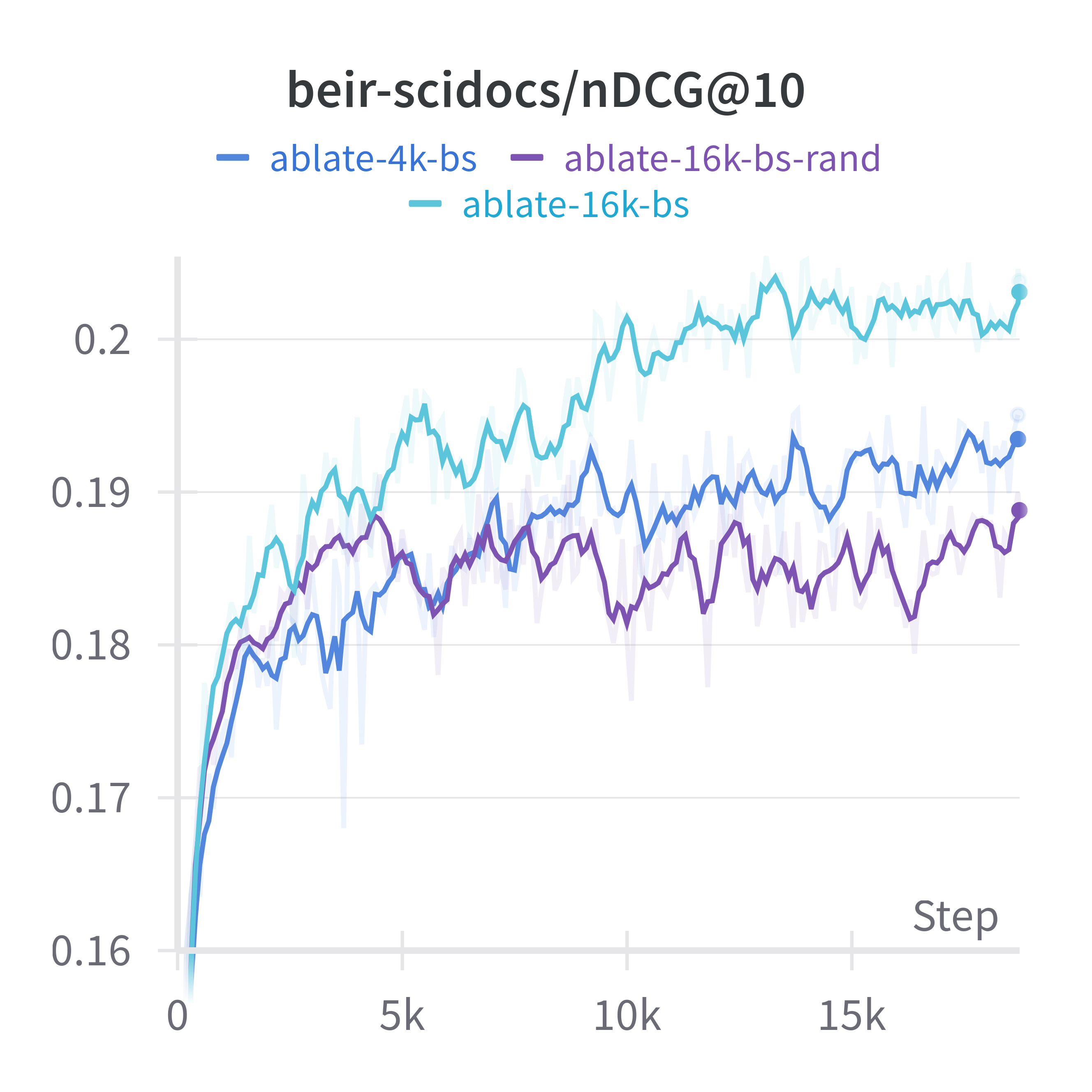}
    \caption{A granular look at our source stratification ablation study (also see Table~\ref{tab:pretrain-ablate}) showing rolling average nDCG@10 on the SciDocs dataset. A large batch size and source stratification (light blue) delivers the highest performance. Although the random-source large batch size run (purple) drives performance up sharply at the start of training, without source stratification, performance quickly plateaus, falling behind the source-stratified small batch run (dark blue) despite using $4\times$ the data and compute.}
    \label{fig:ablate-bs-stratify}
\end{figure}

\begin{table*}
    \centering
    \small
    \begin{tabular}{lllllll}
        \toprule
         Run & Base Model& Data & Stratify Source&  Batch Size&  Seq. Length & Score \\
         \midrule
         A & \texttt{bert-base-uncased} & Snowflake & Yes & 16,384 & 256 & 46.97 \\
         B & \textbf{\texttt{e5-unsupervised-base}} & Snowflake & Yes & 16,384 & 256 & 46.96 \\
         C & \texttt{bert-base-uncased} & \textbf{Nomic} & Yes & 16,384 & 256 & 46.55 \\
         D & \texttt{bert-base-uncased} & Snowflake & \textbf{No} & 16,384 & 256 & 43.74 \\
         E & \texttt{bert-base-uncased} & Snowflake & Yes & \textbf{4,096} & 256 & 45.36 \\
         F & \texttt{bert-base-uncased} & Snowflake & Yes & 16,384 & \textbf{128} & 45.53 \\
         
    \end{tabular}
    \caption{Large-scale training ablation study. Varied treatment bolded. The score is nDCG@10 on the MTEB Retrieval benchmark.}
    \label{tab:pretrain-ablate}
\end{table*}

We probed the effects of batch size, sequence length, base model, and training data in a series of ablations, with resulting MTEB Retrieval scores tabulated in \Cref{tab:pretrain-ablate}. In each case we trained for 20k steps using a linear learning rate decay from 2e-4 to 2e-5 after a 300-step linear warmup from 0.\footnote{This ablation setup is slightly different from our published models' configuration -- the warmup was 300 steps instead of 100, gradient clipping was enabled, and 20k steps often slightly exceeded the one epoch through the data used in our published models (often one epoch was around 19k steps). In some cases, we evaluated a one-epoch checkpoint (around 19k steps instead of 20k) to mitigate a data loading correctness issue discovered post-training for the beyond-one-epoch regime for this dataset.}

Overall, the ablation study results support our hypotheses about data sourcing, longer sequence length, and source stratification improving model performance. In contrast, the choice of initializing from a pre-trained retrieval model did not significantly impact the MTEB Retrieval score after pretraining. We also notice the interesting curriculum-learning-like pattern of source stratification mattering more later in training than other factors like batch size (see \Cref{fig:ablate-bs-stratify}).

\subsection{Fine-tuning Ablations}\label{sec:fine-tune-ablation}

\begin{figure}
    \centering
    \includegraphics[width=1.0\linewidth]{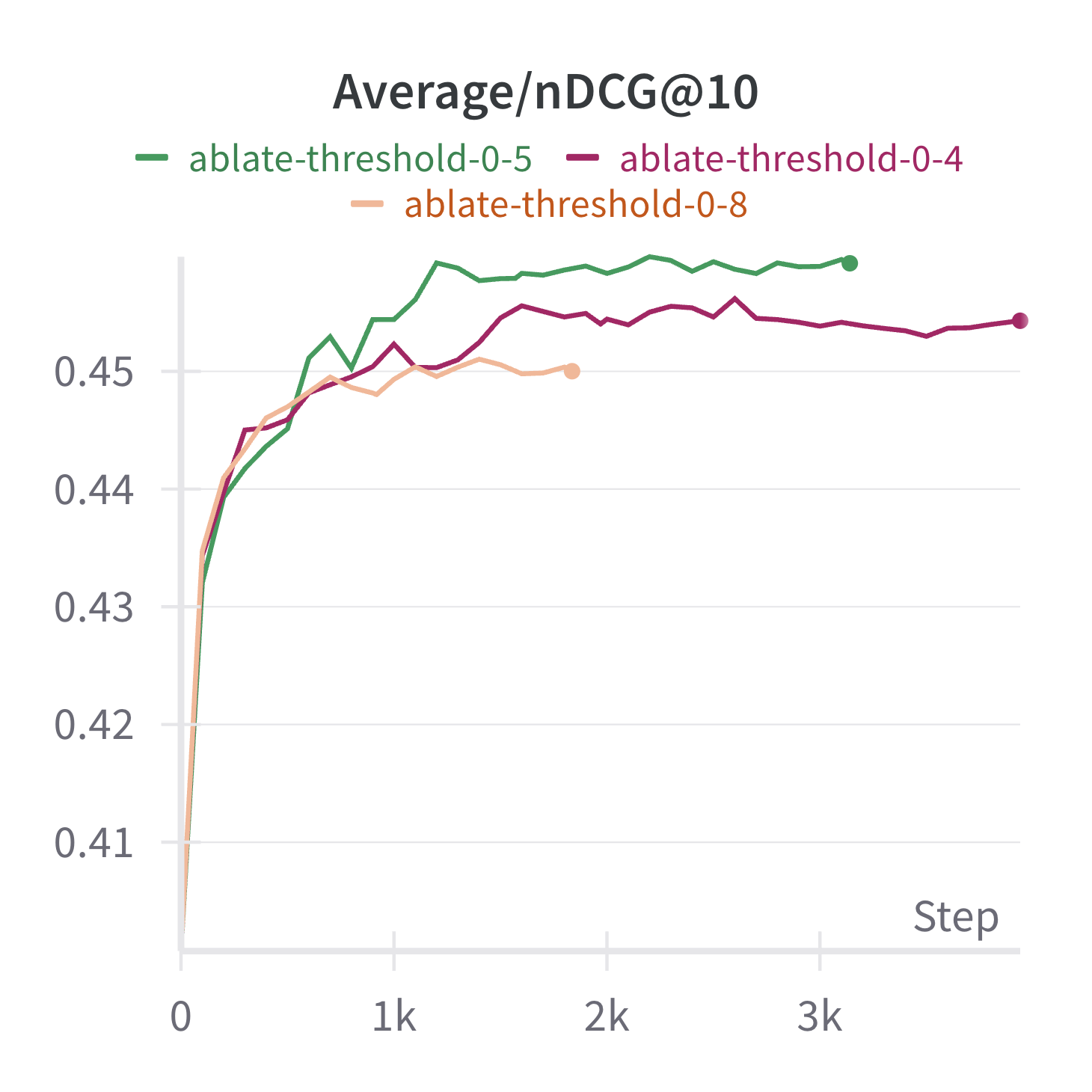}
    \caption{Ablation study for different threshold for hard negatives}
    \label{fig:threshold-hard}
\end{figure}

As discussed in \Cref{sec:optimize}, our tunable negative mining approach uses a threshold to filter out too-hard negatives. We perform an ablation study on several threshold values to demonstrate the importance of the threshold parameter. The results shown in \Cref{fig:threshold-hard} indicate that too-low and too-high maximum relevancy thresholds (too-hard and too-easy negatives) lead to significantly worse performance.

\subsection{End-to-end ablations}\label{sec:ablation-e2e}

\begin{figure}
    \centering
    \includegraphics[width=1.0\linewidth]{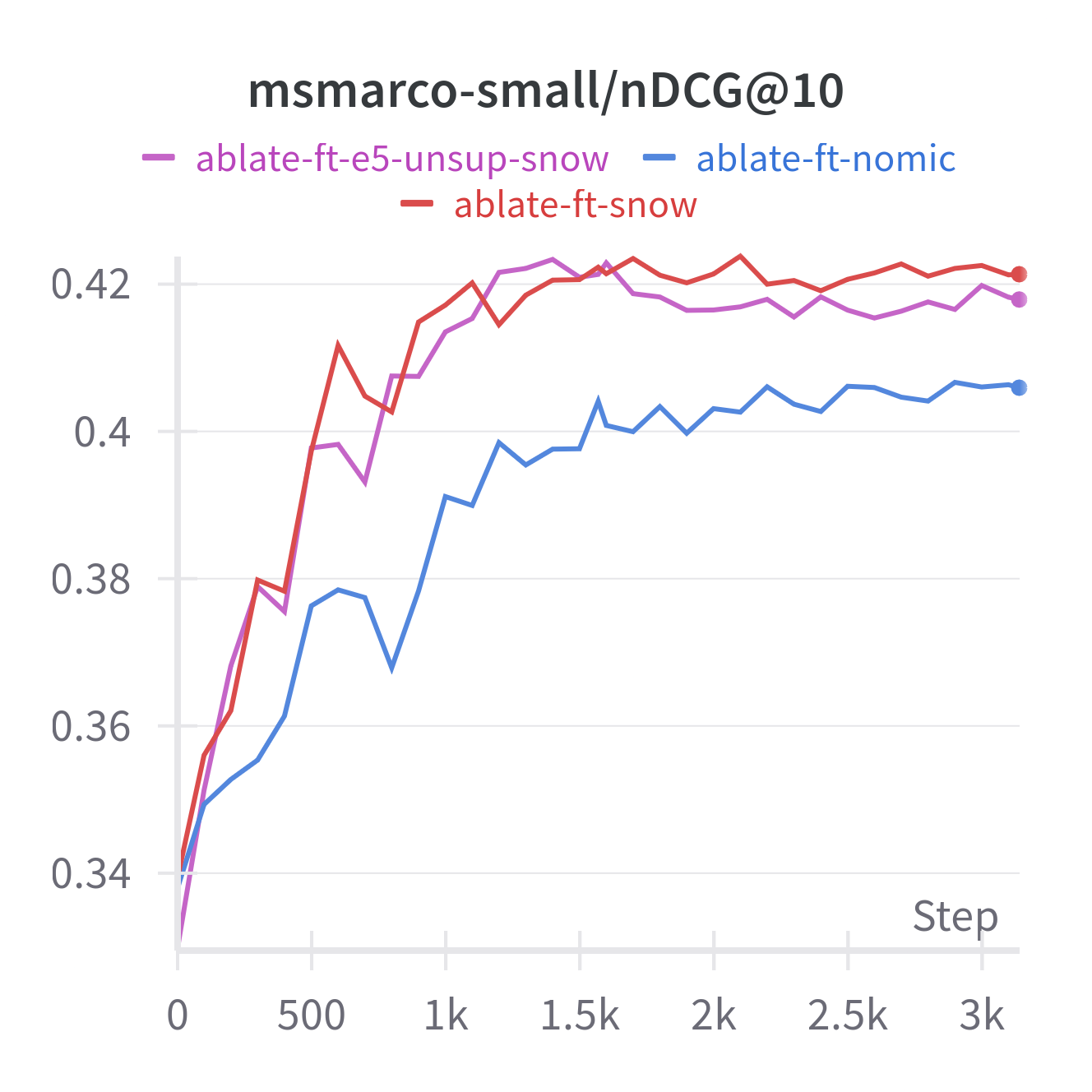}
    \caption{Comparison of various pretrained models under during identical fine-tuning. The performance gap associated with different pretraining datasets appears rather quickly, while the gap associated with different base model weights does not even appear for this dataset (dataset details in \Cref{sec:lite}).}
    \label{fig:ablate-data-ft}
\end{figure}

\begin{table}
    \centering
    \small
    \begin{tabular}{lp{.14\linewidth}p{.19\linewidth}l}
        \toprule
         Starting Model & Pretrain Data & Score After Pretrain & Final Score \\
         \midrule
         \texttt{bert-base} & Snow & 46.97 & 53.92 \\
         \texttt{e5-unsup.} & Snow & 46.96 & 54.67 \\
         \texttt{bert-base} & Nomic & 46.55 & 52.23 \\
         
    \end{tabular}
    \caption{Final two-stage training scores for our published \texttt{m} model two ablations. }
    \label{tab:e2e-ablate}
\end{table}

To thoroughly study the effect of training data on the final score, we extended a subset of our pretraining ablation study through the fine-tuning step. We conducted a finetuning step similar to the one used on our published \texttt{arctic-embed-m} model on configurations A, B, and C from \Cref{tab:pretrain-ablate} (different data and base model). The pretraining and fine-tuning trajectories are shown in \Cref{fig:ablate-data-ft}, with final MTEB Retrieval scores in \Cref{tab:e2e-ablate}. Although the performance gap between models pretrained with Snowflake and Nomic data was relatively modest in pretraining, the gap widens substantially with fine-tuning, despite the fine-tuning recipe is the same. We also see a slight improvement in the final score for the configuration using \texttt{e5-unsupervised-base}. We note that our tuning the fine-tuning step to an \texttt{e5-unsupervised-base} model pre-trained on our data may have affected these results.

\section{Conclusion and Future Work}
By creating the suite of Arctic text embedding models, we sought to better understand how to optimize the training recipe for high-quality text embedding models. Our exploration found that dataset-stratified mini-batches and tuned hard negative mining were crucial ingredients for training a model for more effective retrieval.

In the future, we seek to continue our experimentation to leverage improved curriculum learning and better methods of source stratification. Additionally, we strive to train more robust models to compression approaches such as binarization or quantization of embeddings.

\bibliography{main}
\bibliographystyle{acl_natbib}

\appendix

\section{Dataset Creation Algorithm Details}

Algorithms \ref{algo:negative-mine} and \ref{algo:synthetic} provide the details of the algorithms used to create our high-quality fine-tuning dataset (see \Cref{sec:training-data}).

\begin{table*}
\begin{algoTableCaption}
\caption{Tunable Negative Mining}
\label{algo:negative-mine}
\begin{tabular}{p{\linewidth}}
\hline
\textbf{Require:} 
\begin{itemize}
    \itemsep0em 
    \item $P$, a dataset of $m$ query-document \textit{pair}s 
    \begin{itemize}
        \item $P = \{(q_1, d_1), (q_2, d_2), \ldots, (q_m, d_m)\}$
    \end{itemize}
    \item $D$, a corpus of $n$ documents
    \begin{itemize}
        \item $D = \{d_1, d_2, \ldots, d_n\}$
    \end{itemize}
    \begin{itemize}
        \item One option is for $D$ is to reuse the documents from $P$ 
    \end{itemize}
    \item $r$, a semantic relevance scoring function
    \begin{itemize}
        \item Maps two pieces of text $t_1$ and $t_2$ to a real number, i.e. $\text{score} = r(t_1, t_2)$
        \item Should effectively score the semantic relevance of query-query, document-document, and query-document pairs
        \item Example implementation: use a preexisting text embedding model to embed the query and document, then compute a vector similarity score via cosine similarity.
    \end{itemize}
\end{itemize}

\textbf{Parameters:} 
\begin{itemize}
    \item $R_\text{max}$, a real-valued maximum relevance threshold value
    \begin{itemize}
        \item This threshold defines the degree of relevance below which two items can be considered ``irrelevant'' for the sake of training. This cutoff helps us attenuate label noise and avoid trying to teach the text embedding model to treat relevant documents as irrelevant.
    \end{itemize}
    \item $R_\text{min}$, a real-valued minimum relevance threshold value
    \begin{itemize}
        \item This threshold defines the degree of relevance below which two items are considered ``too obviously irrelevant'' for the sake of training. This cutoff helps us keep all negative examples difficult enough for the model.
    \end{itemize}
    \item $k_\text{neg}$, the maximum desired number of negatives to mine per query-document pair
\end{itemize}
\\ \hline
\vspace{1em}
\begin{enumerate}
    \item Initialize dataset $Z = \{\}$
    \item For each document $d_i$ $\in D$ do 
    \begin{enumerate}
    \itemsep0em 
    \item Compute relevance scores $\pmb{s} = [s_1, \ldots, s_n] = [r(q_i, d_1), r(q_i, d_2), \ldots, r(q_i, d_n)]$
    \item Drop from consideration all documents with relevance scores beyond the minimum or maximum relevance threshold parameters, i.e., define $\pmb{s'} = [s_i: R_\text{min} \leq s_i \leq R_\text{max}]$
    \item Use the scores to determine the most relevant documents $D_\text{topk} = \{d_{i,1}, \ldots, d_{i,k}\} = \topk_{\pmb{s}'}(D)$
    \item Update $Z$ with the query-document-documents example $(q_i, d_i, \{d_{i,1}, \ldots, d_{i,k}\})$, i.e. $Z_\text{updated} = Z \cup \{(q_i, d_i, \{d_{i,1}, \ldots, d_{i,k}\})\}$
    \end{enumerate}
\end{enumerate}    
\\ \hline
\vspace{1em}
\textbf{Variations:}
\begin{itemize}
    \item In addition to positive-query-to-negative-document relevance $r(q_i, d_j)$, positive-document-to-negative-document relevance $r(d_i, d_j)$ can also be used as a signal for mining hard negatives.
\end{itemize}
\\ \hline
\end{tabular}
\end{algoTableCaption}
\end{table*}

\begin{table*}
\begin{algoTableCaption}
\caption{Synthetic Data Generation}
\label{algo:synthetic}
\begin{tabular}{p{\linewidth}}
\hline
\textbf{Require:} 
\begin{itemize}
    \itemsep0em 
    \item $D$, a corpus of $n$ documents (same as Algorithm~\ref{algo:negative-mine})
    \item $g$, a synthetic query generation function
    \begin{itemize}
        \item Maps one positive (relevant) document and $k$ examples of negative (irrelevant) documents to a synthetic query, i.e. $q = g \left(d_\text{pos}, d_\text{neg}^{(1)}, d_\text{neg}^{(2)} \ldots, d_\text{neg}^{(k)}\right)$.
        \item Example implementation: prompt an instruction-tuned LLM using the positive and negative example documents.
    \end{itemize}
    \item $r$, a semantic relevance scoring function  (same as Algorithm~\ref{algo:negative-mine})
\end{itemize}
\\ \hline
\vspace{1em}
\begin{enumerate}
    \item Initialize empty paired dataset $P = \{\}$
    \item For each document $d_i$ $\in D$ do:
    \begin{enumerate}
        \itemsep0em 
        \item Identify the top $k$ documents with highest semantic relevance
        \begin{enumerate}
            \item Compute relevance scores $\pmb{s} = [r(q_i, d_1), r(q_i, d_2), \ldots, r(q_i, d_n)]$
            \item Use the scores to determine the most relevant documents $D_\text{topk} = \{d_{i,1}, \ldots, d_{i,k}\} = \topk_{\pmb{s}}(D)$
        \end{enumerate}
        \item Use generate a synthetic query $q_i = g(d_i, d_{i, 1}, \ldots, d_{i, k})$
        \item Update $P$ with the query pair $(q_i, d_i)$, i.e. $P_\text{updated} = P \cup \{(q_i, d_i)\}$
    \end{enumerate}
    \item Apply Algorithm~\ref{algo:negative-mine} to mine negative documents from paired data $P$ and corpus $D$ using relevance function $r$
\end{enumerate}
\\ \hline
\vspace{1em}
\textbf{Variations:}
\begin{itemize}
    \item In addition to a positive document and a sequence of negative documents, the query generation function can also be constructed to accept an example query to ground the generation. In this case, paired data consisting of query-document pairs becomes required, but higher-quality outputs may be possible in the case of high-quality grounding queries. In this setting, query-document relevance can also be used as a signal for selecting negative examples.
    \item Randomization—rather than taking the top $k$ most relevant documents, one can select the top $k'> k$ documents and randomly sample $k$ of these. This approach (and similar randomization approaches) can generate multiple queries from a single document (or query-document pair if the above variation is used). 
    \item Relevance thresholding -- similar to the machinery of Algorithm~\ref{algo:negative-mine}, one can adjust the relevant negative examples used in query generation by discarding examples deemed ``too relevant'' or ``too irrelevant'' as per thresholds on relevance score.
\end{itemize}
\\ \hline
\end{tabular}
\end{algoTableCaption}
\end{table*}

\begin{table*}
\begin{algoTableCaption}
\small
\caption{Synthetic Data Generation Prompt}
\label{algo:synthetic-query}
\begin{tabular}{p{\linewidth}}
\hline
You are a search quality rater tasked with evaluating the effectiveness of a search engine. You aim to generate a plausible query that retrieves a specific document when executed on a high-performing search engine. The query should be relevant to the document's content and make sense without access to the relevant document. Sample Document: -- BEGIN SAMPLE DOCUMENT -- SAMPLE\_DOC -- END SAMPLE DOCUMENT -- Sample Generated Query: -- BEGIN SAMPLE GENERATED QUERY -- SAMPLE\_QUERY -- END SAMPLE GENERATED QUERY -- Document Details: -- BEGIN DOCUMENT TEXT -- DOCUMENT\_TEXT -- END DOCUMENT TEXT -- Irrelevant Documents: Irrelevant Document 1: WRONG\_1 Irrelevant Document 2: WRONG\_2 Irrelevant Document 3: WRONG\_3 Irrelevant Document 4: WRONG\_4.\\ Instructions: Read the document text and consider potential user actions. Reflect on the document's content and utility. Review the irrelevant documents to understand query nuances. Create a query (Q) that retrieves the target document but excludes irrelevant ones.\\ Considerations: Does the query fully address the user's intent? Does the query uniquely identify the target document? Comprehensive Explanation: Provide a detailed explanation of why the generated query effectively retrieves the target document and excludes irrelevant ones. Ensure the explanation precedes the query for clarity and evaluation purposes.\\ Respond only with the JSON object comprising keys E and Q. \\
\hline 
\end{tabular}
\end{algoTableCaption}
\end{table*}

\section{Additional Efficiency Details}

\Cref{tab:training-efficiency} quantifies our training efficiency in terms of throughput.

\begin{table*}[]
\centering
\begin{tabular}{llllll}
\toprule
Training State                 & Batches & Doc/Batch & Doc Max Length & Elapsed & Doc/Sec \\
\midrule
Large Scale In-Batch Negatives &   18,798 &  16,384  & 256 &   17h3m & 5,018 \\
Smaller Scale Hard Negatives  &    7,845 &   5,632  & 512 &   7h40m &  1,601  
\end{tabular}
\caption{Training efficiency for \texttt{arctic-embed-m} on 8 NVIDIA H100 GPUs. The in-training evaluation is included in the runtime.}
\label{tab:training-efficiency}
\end{table*}

\subsection{Fast Feedback With ``Lite'' Datasets}\label{sec:lite}

We found it valuable to conduct information retrieval evaluation at regular intervals throughout the training process, not just at the end, as this uncovered significant trends not captured by end-of-training evaluation (see \Cref{fig:ablate-bs-stratify}, for example). To enable these evaluations, we constructed ``lite'' versions of several large BEIR datasets by starting with a sample of queries (e.g., a few hundred) and combining the labeled-as-relevant with the most relevant documents as determined using a preexisting text embedding model (we often used 100 documents per query) to form a corpus far smaller than the original yet still difficult given the query test set. We found that these ``lite'' datasets offered a cheap way to anticipate full-scale performance trends at a small fraction of the compute cost required by the entire dataset.

We could evaluate retrieval performance on a diverse set of domains throughout training through a combination of already-small BEIR datasets and our new ``lite BEIR'' datasets. We achieved excellent throughput by implementing evaluation in the same Distributed Data-Parallel paradigm as our training. We were able to embed and score \textasciitilde five datasets of this size in \textasciitilde 30 seconds, netting us useful nDCG@10 scores as often as every 100 steps of training with only modest runtime overhead (\Cref{fig:ablate-bs-stratify,fig:base-training-curve,fig:ablate-schedule}, which are screenshots of our in-training evaluation, demonstrate just how granular and proper this evaluation frequency is in practice).

\subsection{Efficiency Tricks}\label{sec:appendix-efficiency-tricks}

\subsubsection*{Training Efficiency Tricks}
\begin{itemize}
    \item Automatic Mixed Precision\footnote{\url{https://pytorch.org/docs/stable/amp.html}} offers \texttt{bfloat16} training speeds (often a \textasciitilde 2x speedup) with little to no quality degradation compared to \texttt{float32} training. 
    \item PyTorch Distributed Data-Parallel (DDP) includes a \texttt{ZeroRedundancyOptimizer} implementation\footnote{\url{https://pytorch.org/tutorials/recipes/zero_redundancy_optimizer.html}} which can reduce memory consumption, though the savings are modest for typical BERT-scale models.
    \item Enabling PyTorch's \texttt{roundup\_power2\_divisions} and/or \texttt{expandable\_segments} CUDA memory allocator parameters\footnote{\url{https://pytorch.org/docs/stable/notes/cuda.html}} can mitigate memory fragmentation issues arising from inconsistent batch size and sequence length (e.g. from batch deduplication or padding each batch to the length of the longest sequence it contains). 
    \item The DisCo trick for distributed contrastive loss \cite{chen2023discoclip} is just a few extra lines of PyTorch code\footnote{\url{https://github.com/IDEA-Research/DisCo-CLIP/blob/main/disco/gather.py}}.
\end{itemize}

\section{List of Data Quality Filter Heuristics}\label{sec:appendix-filter-list}

\begin{itemize}
    \item \textbf{Language}: Only include documents primarily identified as English using the fasttext language classifier threshold.
    \item \textbf{Document Length}: The number of normalized words in each document should be between 10 and 10,000. This filter removes very short or extremely long documents.
    \item \textbf{Word Length}: After normalization, the mean length of words should be between 3 and 10 characters. This filter helps remove documents with excessively short or long words.
    \item \textbf{Symbol Density}: The document's ratio of symbols to words should be less than 0.1 (10\%). This filter removes documents with an excessive number of symbols or special characters.
    \item \textbf{Ellipsis Line}: The fraction of lines that end with an ellipsis (...) should be less than 0.3 (30\%). This filter removes documents with an excessive number of lines ending with ellipses.
    \item \textbf{Non-Alphabetical Word}: The fraction of words that contain no alphabetical characters should be less than 0.2 (20\%). This filter removes documents with excessive non-alphabetical words (e.g., numeric strings, symbols).
    \item \textbf{Perplexity}: The perplexity score (from \url{https://github.com/kpu/kenlm}) should be less than 10,000. This filter removes documents that are less likely to be understandable or meaningful.
    \item \textbf{N-Gram Duplication}: Limit the fraction of characters in duplicate n-grams (sequences of n words) for n ranging from 2 to 10. These filters help remove documents with excessive repetition or duplication of phrases.
    \item \textbf{Stop Word}: The document should contain at least one stop word (common words like ``the'', ``and'', ``is''). This filter helps remove documents without meaningful content.
    \item \textbf{Bullet Point Line}: The density of lines starting with bullet points should be less than 0.9 (90\%). This filter removes documents that are mostly composed of bullet points or lists.
    \item \textbf{Blacklist Domain}: Domain of document URL cannot be blacklisted based on the \url{https://dsi.ut-capitole.fr/blacklists/index_en.php}. This filter helps remove documents from known low-quality, adult, or spam domains.
    \item \textbf{Short Line}: The fraction of lines with fewer than five words should be less than 0.1 (10\%). This filter removes documents with an excessive number of concise lines.
    \item \textbf{Numeric Line}: The fraction of lines containing only numeric characters should be less than 0.05 (5\%). This filter removes documents with excessive lines consisting solely of numbers.
    \item \textbf{Uppercase Line}: The fraction of lines with more than 80\% uppercase letters should be less than 0.05 (5\%). This filter removes documents with excessive lines in all uppercase letters.
\end{itemize}

\section{Anecdata On The Importance Of Tuning}\label{sec:appendix-tuning}
Throughout the development of Arctic Embed, we found the time-honored technique of guess-and-check tuning to be critical for ensuring a quality final model. By investing in efficient training and high-granularity in-training evaluation (see \Cref{sec:efficiency}), we were able to carry out dozens of experiments ranging from one-off wild guesses (``YOLO runs'') to large-scale parameter sweeps spanning ten or more trials in the background of our work to build our in-house datasets and finish our training code implementation.

\begin{figure}
    \centering
    \subfloat[nDCG@10 score averaged across several datasets]{%
      \includegraphics[clip,width=\columnwidth]{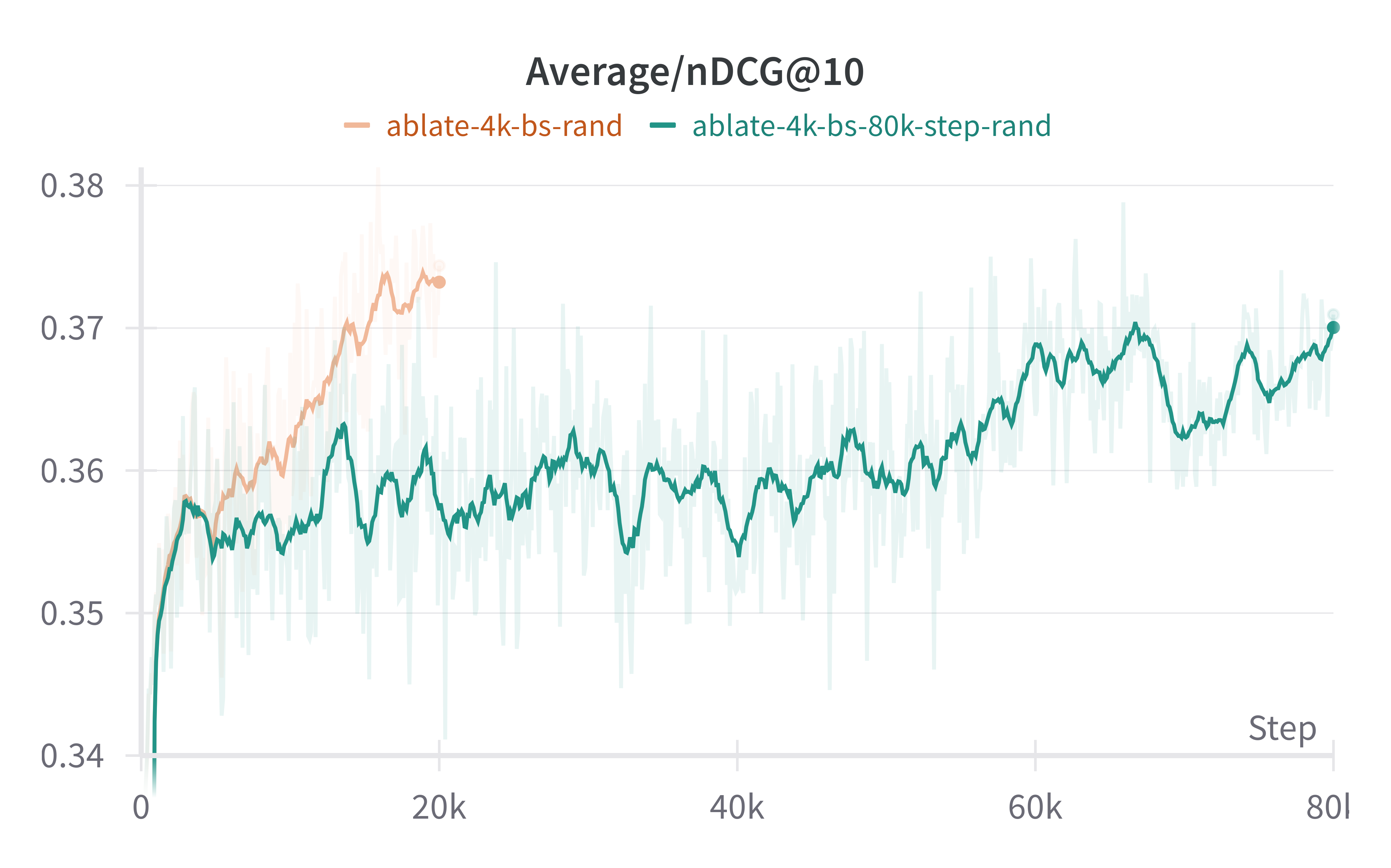}\label{subfig:score}%
    }
    
    \subfloat[Loss curves]{%
      \includegraphics[clip,width=\columnwidth]{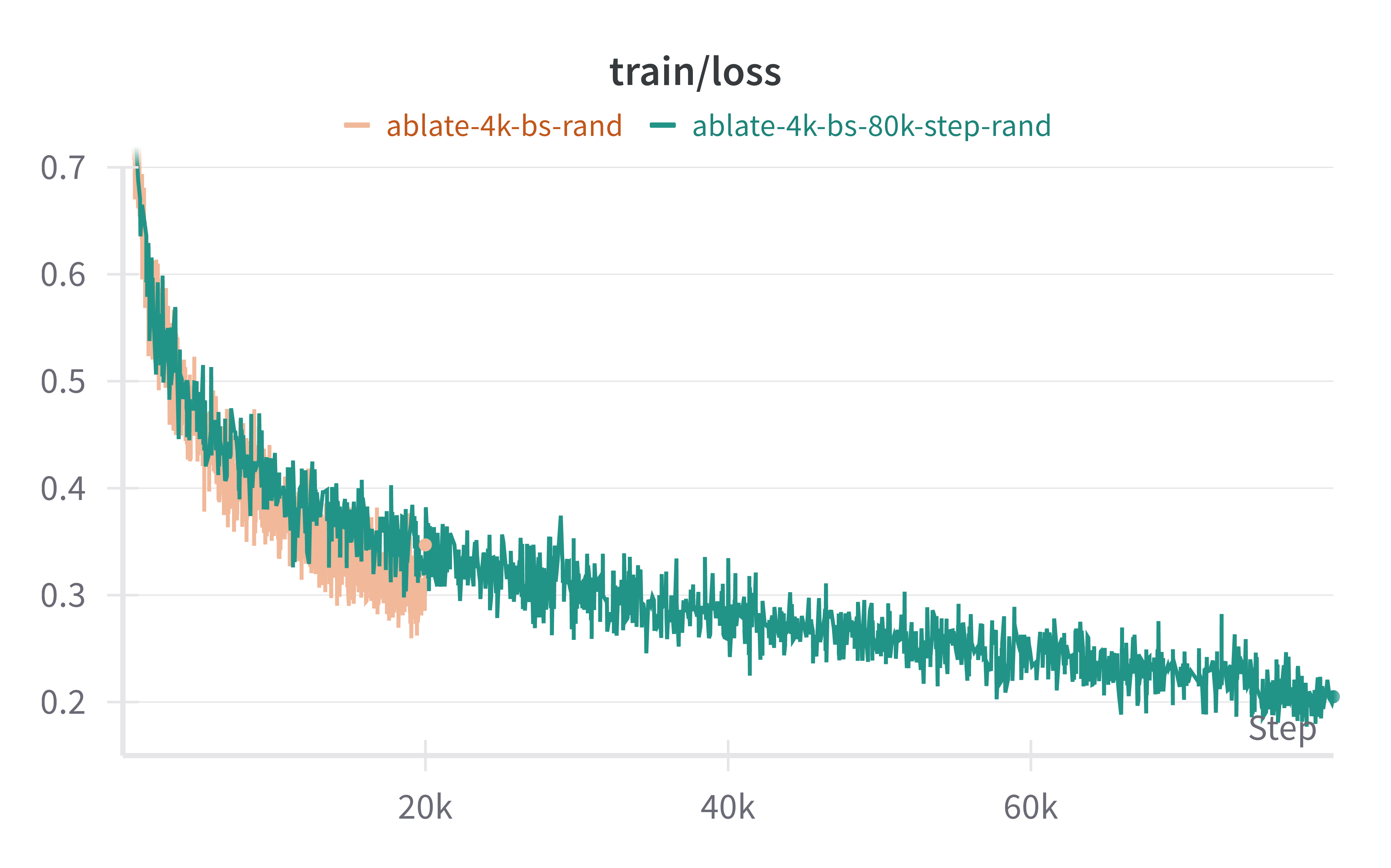}\label{subfig:loss}%
    }
    \caption{Example of surprising sensitivity to learning rate schedule. The yellow lines track a 20k step training run, while the green lines show the same data and hyperparameters extended to an 80k step run, with the only difference in the first steps being the slower linear learning rate decay in the longer run. Although learning rates are still similar at 6k steps (approximately 0.00015 and 0.00019 for yellow and green, respectively), the learning trajectories diverge sharply around this point, with the 20k schedule learning faster both in terms of downstream IR performance (Figure~\ref{subfig:score}) and in-sample contrastive loss (Figure~\ref{subfig:loss}). }
    \label{fig:ablate-schedule}
\end{figure}

\begin{table}
    \centering
    \small
    \begin{tabular}{lll}
        \toprule
         Stratify Source& Steps & Score \\
         \midrule
         Yes & 20k & 45.36 \\
         Yes & 80k & 46.13 \\
         No & 20k & 42.42 \\
         No & 80k & 42.79 \\        
    \end{tabular}
    \caption{Extended study of 4k batch size runs demonstrating better performance at fewer steps. A one-epoch (~75k step) checkpoint is used in the 80k stratified due to a data loading bug.}
    \label{tab:4k-ablate}
\end{table}

\begin{figure}
    \centering
    \includegraphics[width=1.0\linewidth]{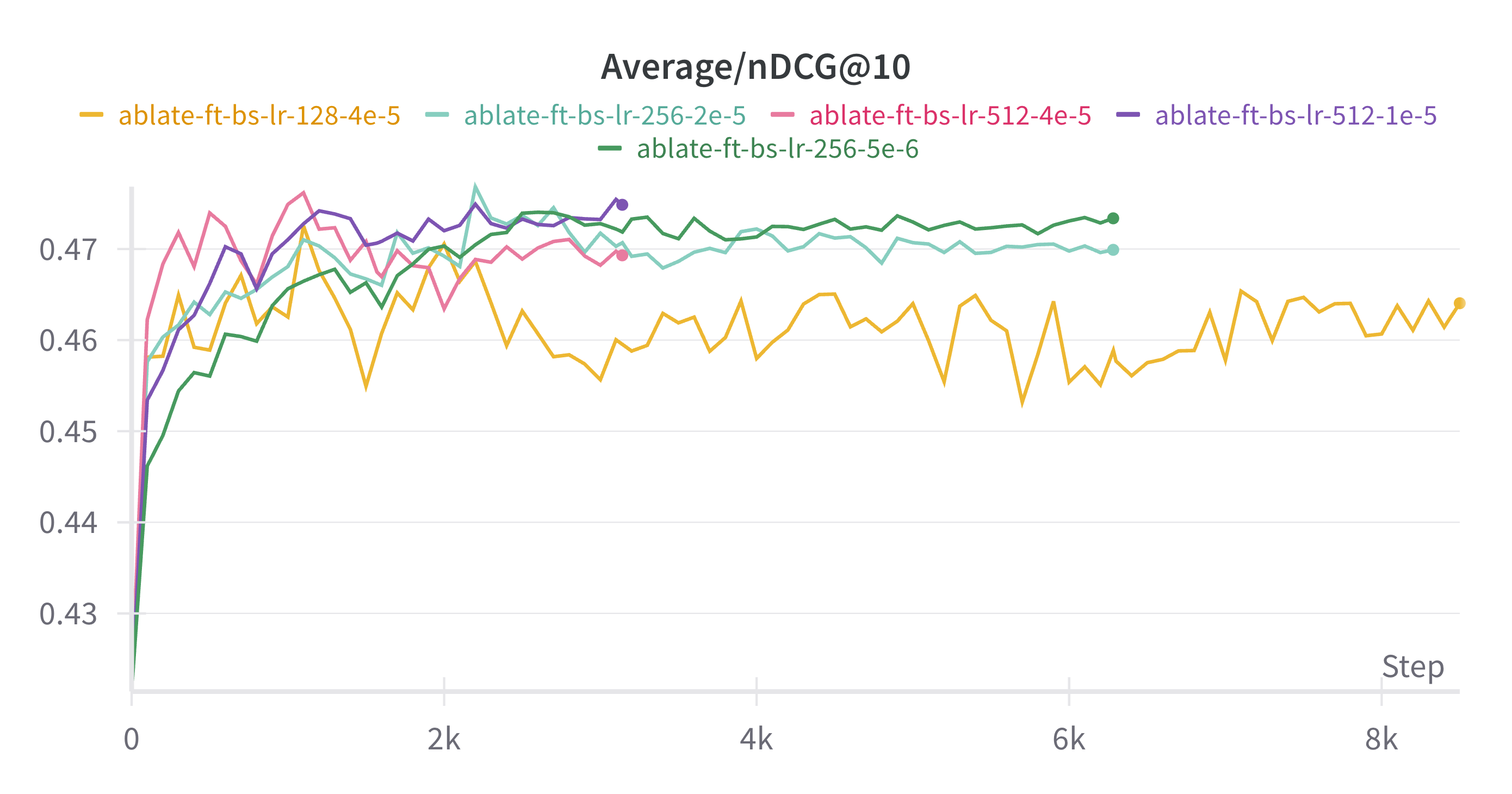}
    \caption{Hyperparameter sweep for different batch sizes and learning rates.}
    \label{fig:ft-lr-bs}
\end{figure}

We were surprised quite often in our tuning. Sometimes, we were surprised by an unexpected insensitivity to presumably important hyper-parameters, as was the case when we experimented with various batch size settings and learning rates for fine-tuning as shown in \Cref{fig:ft-lr-bs}. Other times, we were surprised by unexpected sensitivity, like the unforeseen differences in trajectory between shorter and longer training runs during (see the comparisons in \Cref{tab:4k-ablate} and \Cref{fig:ablate-schedule} for example -- where we see that training for more steps with our linear learning rate decay schedule leads to much worse performance at the beginning of training).

Although we ran many informal experiments while developing Arctic embed, we believe there is still much to be learned about the peculiarities of tuning text embedding training.
\section{Full MTEB Score Breakdown}\label{sec:appendix-full-table}

\newgeometry{margin=1cm} % small margin
\begin{landscape}
\begin{table}
\centering
\small

\scalebox{0.9}{\begin{tabular}{
lp{0.08\textwidth}p{0.08\textwidth}p{0.08\textwidth}p{0.08\textwidth}p{0.08\textwidth}p{0.08\textwidth}p{0.08\textwidth}p{0.08\textwidth}p{0.08\textwidth}p{0.08\textwidth}p{0.08\textwidth}p{0.08\textwidth}
}

\toprule
 & baseline & using e5-base-unsupervised & using nomic data & using sequence length 128 & 4k batch 80k step stratified & 4k batch 80k step unstratified & 4k batch 20k step stratified & 4k batch 20k step unstratified & without source stratification & baseline, finetuned & using e5-base-unsupervised, finetuned & using nomic data, finetuned \\
\midrule
ArguAna & 45.14 & 43.95 & 53.66 & 39.69 & 44.13 & 39.70 & 44.75 & 42.18 & 45.65 & 58.77 & 55.44 & 51.78 \\
CQADupstackRetrieval & 42.26 & 42.24 & 41.57 & 40.80 & 40.71 & 37.36 & 40.15 & 36.90 & 39.05 & 43.24 & 44.11 & 42.75 \\
ClimateFEVER & 20.35 & 23.75 & 22.09 & 21.33 & 20.73 & 19.51 & 18.47 & 19.78 & 18.79 & 35.55 & 38.82 & 32.57 \\
DBPedia & 38.97 & 37.83 & 38.35 & 36.92 & 38.77 & 37.58 & 39.27 & 35.77 & 37.02 & 44.68 & 45.81 & 43.07 \\
FEVER & 71.65 & 71.39 & 69.59 & 64.42 & 70.61 & 71.04 & 72.55 & 68.34 & 68.82 & 87.55 & 88.46 & 86.78 \\
FiQA2018 & 42.09 & 45.20 & 37.40 & 40.46 & 39.70 & 35.30 & 38.73 & 34.92 & 37.67 & 40.44 & 42.88 & 36.79 \\
HotpotQA & 57.77 & 53.97 & 60.52 & 56.69 & 59.10 & 47.22 & 60.54 & 47.22 & 46.27 & 71.77 & 73.03 & 69.84 \\
MSMARCO & 33.85 & 33.75 & 34.35 & 33.77 & 33.12 & 31.46 & 32.79 & 31.05 & 32.40 & 41.73 & 41.89 & 40.26 \\
NFCorpus & 33.83 & 36.28 & 36.25 & 32.83 & 34.28 & 32.71 & 32.81 & 31.90 & 32.58 & 34.63 & 36.16 & 34.98 \\
NQ & 46.92 & 47.07 & 48.46 & 46.68 & 45.81 & 37.30 & 44.87 & 37.55 & 38.85 & 60.37 & 61.82 & 58.48 \\
QuoraRetrieval & 87.20 & 87.45 & 88.13 & 87.30 & 86.99 & 86.25 & 86.51 & 86.10 & 86.58 & 87.32 & 87.60 & 87.94 \\
SCIDOCS & 20.39 & 21.54 & 21.62 & 20.24 & 20.02 & 17.88 & 19.09 & 18.30 & 18.58 & 20.20 & 21.09 & 21.00 \\
SciFact & 69.42 & 70.97 & 74.97 & 68.90 & 68.68 & 65.99 & 66.89 & 66.88 & 68.36 & 70.58 & 73.34 & 75.16 \\
TRECCOVID & 71.32 & 66.99 & 50.91 & 70.96 & 66.82 & 62.14 & 61.23 & 59.14 & 63.51 & 79.98 & 80.27 & 74.70 \\
Touche2020 & 23.38 & 21.97 & 20.36 & 22.00 & 22.42 & 20.34 & 21.76 & 20.22 & 22.03 & 31.95 & 29.38 & 27.40 \\
\midrule
Overall Average & 46.97 & 46.96 & 46.55 & 45.53 & 46.13 & 42.79 & 45.36 & 42.42 & 43.74 & 53.92 & 54.67 & 52.23 \\
\bottomrule
\end{tabular}}

\caption{Full Per Dataset NDCG@10 MTEB Retrieval scores by dataset for each ablation variant. }
\end{table}
\end{landscape}
\restoregeometry

\end{document}